\documentclass{bmvc2k}

\title{\hspace{15pt}Slim-CNN: A Light-Weight CNN for Face\\ \hspace{90pt}Attribute Prediction}

\addauthor{Ankit Sharma}{ankit.sharma285@knights.ucf.edu}{1}
\addauthor{Hassan Foroosh}{foroosh@cs.ucf.edu}{1}

\addinstitution{
 Department of Computer Science\\
 University of Central Florida\\
 Florida, USA
}

\runninghead{Slim-CNN}{Light-Weight CNN}

\begin{document}

\maketitle

\begin{abstract}
We introduce a computationally-efficient CNN micro-architecture \textit{Slim Module} to design a lightweight deep neural network \textit{Slim-Net} for face attribute prediction. Slim Modules are constructed by assembling depthwise separable convolutions with pointwise convolution to produce a computationally efficient module. The problem of facial attribute prediction is challenging because of the large variations in pose, background, illumination, and dataset imbalance. We stack these Slim Modules to devise a compact CNN which still maintains very high accuracy. Additionally, the neural network has a very low memory footprint which makes it suitable for mobile and embedded applications. Experiments on the CelebA dataset show that Slim-Net achieves an accuracy of 91.24\% with at least 25 times fewer parameters than comparably performing methods, which reduces the memory storage requirement of Slim-net by at least 87\%.   

\end{abstract}

\section{Introduction}
\label{sec:intro}
The introduction of Convolutional Neural Networks (CNN) has led to the advancement of state-of-the-art performance on several fundamental computer vision tasks such as image classification \cite{szegedy}, semantic segmentation \cite{DeepLab}, and object detection \cite{SSD}. The success of popular CNN architectures such as VGG-Net \cite{VGG} and Res-Net \cite{he} has steered the research into building deeper and more parameter intensive networks. While these networks yield higher accuracy than their predecessors, they are not very economical. This is particularly important when considering deployment on mobile-powered platforms, where the storage burden and computational cost of parameter intensive networks are most noticeable. In recent years, a few efficient architectures, like Mobile-Net \cite{MobileNet} and Shuffle-Net \cite{ShuffleNet}, have been proposed to address these concerns. Their key idea is to design a deep network using computationally economical types of convolution such as depthwise separable convolution and group convolutions. \vspace{3 pt}

\begin{figure}[t]
\hspace{25pt}
\includegraphics[width=0.9\linewidth]{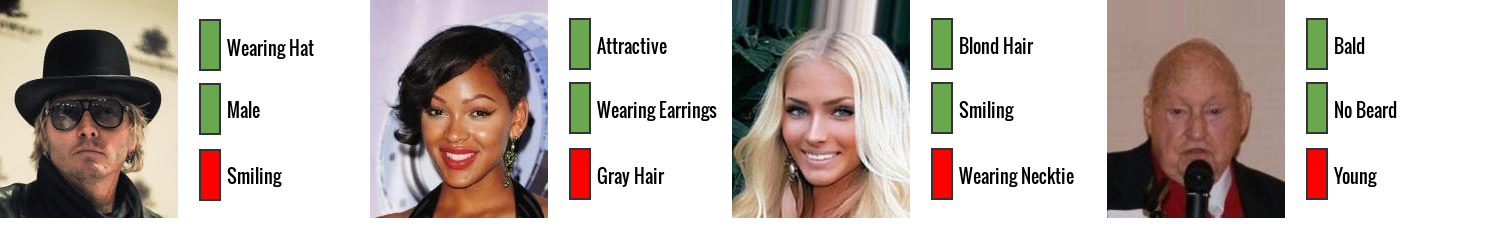}
\caption{Facial Attribute Prediction: Given an image, the task is to detect the presence or absence of attributes. Consider the figure shown above. If an attribute is present, the corresponding box is colored green otherwise it is colored red. Images are taken from CelebA dataset \cite{liu}.}
\label{fig:AttributeDefinition}
\end{figure}

Analysis of facial attributes provides a very important visual cue for a range of applications like surveillance, biometric recognition systems, retrieval, fraud detection, etc. Over the last few years, there has been significant interest in the multi-label face attributes problem as shown in Fig.\ref{fig:AttributeDefinition}. Most of the previous work on facial analysis have used Alex-Net \cite{AlexNet} inspired architectures that made the final network architecture more computationally expensive and parameter exhaustive. This makes it unsuitable for mobile or embedded applications where face attribute prediction is used as one of the main building blocks. \vspace{3 pt}

The aim of our work is to design a computationally efficient, high performance CNN architecture that is lighter parameter wise than existing architectures. To achieve this goal we focus on the use of CNN micro-architectures, that is the arrangement of convolutional, pooling, fully-connected, and dropout layers into individual modules. Micro-architectures, like Inception module\cite{szegedy} and Residual blocks\cite{he}, have been shown to give good discriminative feature representations. In this paper, we propose to use our novel micro-architecture \textit{Slim Module} to build an efficient deep neural network. The \textit{Slim Module} is composed of convolutional layers of two different types (depth separable convolution for 3x3 kernel and pointwise convolution for 1x1 kernel) along with skip-connections. \vspace{3 pt}

To summarize, our main contributions are as follows: 
\begin{itemize}
\item We introduce a novel CNN micro-architecture \textit{Slim Module} that is compact, computationally efficient, has a smaller memory footprint, and still yields very good discriminative features.
\item We thoroughly investigate our proposed micro-architecture for different values of kernel sizes and compare it with other micro-architectures in the literature. 
\item We compare the performance of a stacked \textit{Slim Module} deep neural network \textbf{Slim-CNN} with state-of-the-art methods on face attribute prediction in terms of accuracy, parameter size, and on-disk memory space. 
\end{itemize}

\section{Related Work}
We will review two lines of related work: face attribute prediction and micro-architectures in CNNs.

\subsection{Face Attribute Prediction}
Kumar, et al. \cite{kumar} were the first to introduce the classification of facial attributes for the task of face verification. They used binary classifiers to detect the presence or absence of attributes which they called \textit{attribute classifiers}. Liu, et al. \cite{liu} introduced the large scale CelebFaces Attribute (CelebA) Dataset for prediction of facial attributes in the wild. They proposed a deep learning framework of cascading DNNs consisting of two localization networks (LNet) and an attribute recognition network (ANet) for face localization and face attribute classification. Zhong, et al. \cite{zhong} used mid-level CNN features to construct deep hierarchical feature representations for facial attribute prediction. The authors argue that some attributes are more locally-defined and earlier layers of CNNs are more suitable for the identification of these attributes. In \cite{moon}, Rudd uses a multi-task approach to solve the problem of attribute classification. In order to deal with the multi-label imbalance problem in the dataset, they introduce mixed objective optimization network(MOON). Since the learning of each facial attribute is treated as a different task, the cost function is set-up for joint optimization.

Hand \cite{hand} leveraged the relationships between attributes to build a multi-task deep convolutional network (MCNN) wherein they manually divided 40 attributes into 9 groups according to each attribute's location. The lower layers of MCNN are shared among all the groups while the higher layers are shared only for related attributes. They attached a fully-connected layer to the output of the MCNN (MCNN-AUX) to identify relationships between different attribute scores and to improve the final accuracy. Han et al. \cite{han} used a multi-task approach to estimate facial attributes by modeling both attribute correlation and attribute heterogeneity in a single network. The authors initialized the network by training it for face verification on the CASIA-Webface dataset \cite{Casia}. Gunther \cite{Affact} suggested using data augmentation to learn facial attributes on a ResNet architecture without the need of any sort of facial alignment. Note that all of the mentioned works on facial attribute prediction have focused primarily on performance without addressing the need for compact and efficient models.

\begin{figure}[t]
\centering
\includegraphics[width=0.8\linewidth]{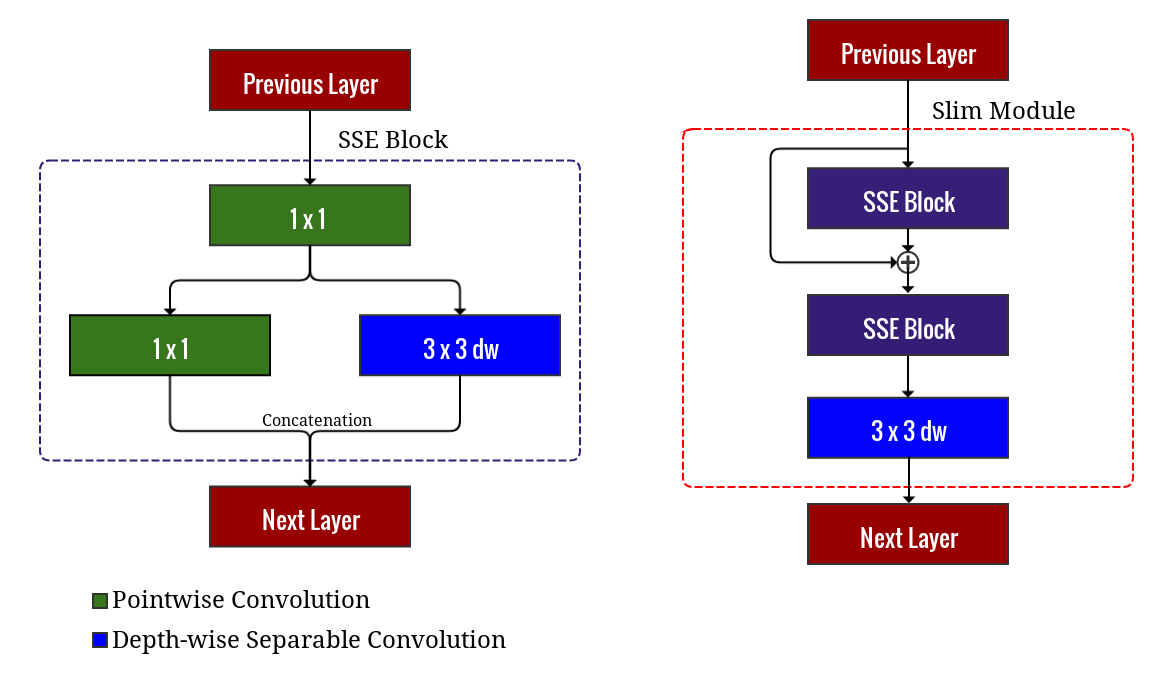}
\caption{Slim Module: The micro-architecture shown on the left is the Separable Squeeze-Expand Block (SSE), while on the right is the \textit{Slim Module}.} 
\label{fig:MA}
\end{figure}

\subsection{Micro-Architecture in CNN}
Micro-architecture refers to the organization of individual layers and modules. Over the last few years, micro-architectures such as Inception Modules and Residual Modules have become very popular as building blocks for constructing deep neural networks. Lin \cite{nin} proposed a DNN "Network In Network" which consists of stacking multiple micro neural network structures "mlpconv layer" to enhance the abstraction ability of the model. In \cite{szegedy}, Szegedy stacked these Inception Modules to get superior performance while still keeping the computational complexity in check. They implemented dimension reduction by using 1x1 convolution kernels in order to lower the number of parameters. Landola  \cite{SqueezeNet} presented a compact Fire Module to design a small deep network to get same performance as Alex-Net, but with 50 times fewer parameters. In  \cite{he}, He introduced a residual learning framework for training DNNs with very large depths without sustaining a loss of performance. The idea behind He's work is to fit a residual mapping with a reference to the input rather than to the desired underlying mapping. Several works \cite{mask, xie} have since used this principle to build smaller residual modules for training different tasks.

\section{Method}
Our goal is to design a new CNN micro-architecture to be used as the building blocks of our deep neural network. There has been extensive research into neural network architectures, which has led to many structural innovations. These innovations have contributed significantly toward improving the performance of CNNs across many different computer vision tasks. As such, they were considered when designing our micro-architecture. The following structural innovations have been incorporated into our \textit{Slim Module}: \vspace{5 pt}

1. \textbf{Multiple Branches:} From the success of Inception and ResNets, it can be inferred that having multiple branches in the network provides better feature representations. Multiple branches in a network means multiple paths through the network which would lead to good discriminative features. \vspace{5 pt}

2. \textbf{Small Kernels:} VGG-Net \cite{VGG} and GoogleNet \cite{szegedy} showed that smaller kernels such as [1x1, 3x3, 5x5] can improve the performance of a deep neural network while also reducing the number of parameters of the network. Smaller kernels have the added advantage of being more computationally efficient. Also, stacking two 3x3 convolutional layers have effectively the same receptive field as that of 5x5 convolutional layer. This means that combination of smaller kernels can simulate a larger kernel while keeping the number of parameters significantly lower. \vspace{5 pt}

3. \textbf{Skip-Connections:} Skip-Connections \cite{highway,he}, also called by-pass connections, are defined as the connections between nodes in different layers of a neural network that skip one or more layers of processing. Skip-connections allow neural networks to regulate information by providing paths along which information can go through several layers without any attenuation. This has been highlighted by the success of architectures that utilize skip-connections like in ResNets and HighWay \cite{highway} Networks.

\begin{table}[t]
\footnotesize
\centering
\caption{Configuration Details of the proposed Slim-Net} \vspace{1pt}
\label{SlimNetTable}
\begin{tabular}{cccccc}
\hline
\textbf{\begin{tabular}[c]{@{}c@{}}Layer\\ Name\end{tabular}}  & \textbf{\begin{tabular}[c]{@{}c@{}}Filter Size/\\ Stride\end{tabular}} & \textbf{Squeeze} & \textbf{\begin{tabular}[c]{@{}c@{}}Expand\\ (1x1)\end{tabular}} & \textbf{\begin{tabular}[c]{@{}c@{}}Expand\\ (3x3)\end{tabular}} & \textbf{\begin{tabular}[c]{@{}c@{}}DW Separable\\ (3x3)\end{tabular}} \\ \hline
\multicolumn{6}{c}{Input Layer of size 178 x 218 x 3}                                                                                                                              \\ \hline
\begin{tabular}[c]{@{}c@{}}Conv2D \\ (96 filters)\end{tabular} & 7x7/ 2x2                                                               &                  &                                                                 &                                                                 &                                                                       \\ \hline
MaxPool                                                        & 3x3/2x2                                                                &                  &                                                                 &                                                                 &                                                                       \\ \hline
Slim Module                                                    &                                                                        & 16               & 64                                                              & 64                                                              & 48                                                                    \\ \hline
MaxPool                                                        & 3x3/2x2                                                                &                  &                                                                 &                                                                 &                                                                       \\ \hline
Slim Module                                                    &                                                                        & 32               & 128                                                             & 128                                                             & 96                                                                    \\ \hline
MaxPool                                                        & 3x3/2x2                                                                &                  &                                                                 &                                                                 &                                                                       \\ \hline
Slim Module                                                    &                                                                        & 48               & 192                                                             & 192                                                             & 144                                                                   \\ \hline
MaxPool                                                        & 3x3/2x2                                                                &                  &                                                                 &                                                                 &                                                                       \\ \hline
Slim Module                                                    &                                                                        & 64               & 256                                                             & 256                                                             & 192                                                                   \\ \hline
MaxPool                                                        & 3x3/2x2                                                                &                  &                                                                 &                                                                 &                                                                       \\ \hline
\multicolumn{6}{c}{Global Average Pooling}                                                                                                                                                                                                                                                                                                                           \\ \hline
Fully-Connected  &   40   &     &     &    &    \\ \hline 
\end{tabular}
\end{table}

\vspace{5pt}
\subsection{Proposed Micro-Architecture: } 
CNN micro-architecture, as higher level building blocks, present a modular approach to designing deep neural networks by reducing the need to manually choose the type and filter dimensions of each convolutional layer. The proposed Slim module, shown in Figure \ref{fig:MA}, is made up of Separable Squeeze-Expand (SSE) blocks and depthwise separable convolutional layers. Depthwise separable convolutions consists of depthwise convolution, a spatial convolution performed independently over each channel yielding a new channel space, followed by pointwise convolution, a 1x1 convolution over the channel-space output of the depthwise convolution. These convolutions have fewer parameters than the standard convolutions which make them less prone to overfitting. Additionally, these convolutions are computationally cheaper and faster than the standard convolutions which make them a great fit for mobile and embedded applications.

The SSE block is a multi-layered arrangement consisting of two 1x1 pointwise convolutional layers and a single 3x3 depthwise separable convolutional layer. The first layer, called the "Squeeze" layer, is a 1x1 convolutional layer with fewer filter dimensions than the previous layer which compresses the feature representation. The next layer is a concatenation of a 1x1 pointwise convolutional filter and a 3x3 depthwise separable convolution filter. This is the "Expand" layer as it significantly widens the number of output channels, increasing the filter dimensions to four times that of the squeeze layer. The SSE block is similar to the fire module described in the paper\cite{SqueezeNet}, but we improve on it by using depthwise separable convolution. 

The \textit{Slim Module} stacks two SSE blocks together followed by a 3x3 depthwise separable convolution layer. Additionally, there is a skip-connection over the first SSE block, resulting in the sum of the input and output of the first block being used as the input to the second SSE block. In our micro-architecture design, the convolutional layers are followed by a batch normalization layer and a ReLu activation layer.

In addition to making the \textit{Slim Module} cost-effective, we sought to make it easy to configure. To that end, we define a hyperparameter for the \textit{Slim Module} called \textit{FilterCount} which is the number of filters in the squeeze layer of the SSE Blocks for a given \textit{Slim Module}. The number of filters for the other convolutional kernels in that module are just multiples of the \textit{FilterCount}. To get the number of convolutional filters in the Expand layer and the 3x3 depthwise separable convolutional layer after the second SSE Block, the \textit{FilterCount} was multiplied by 4 and 3, respectively.

\begin{figure*}[t]
\centering
 \includegraphics[width=0.8\linewidth]{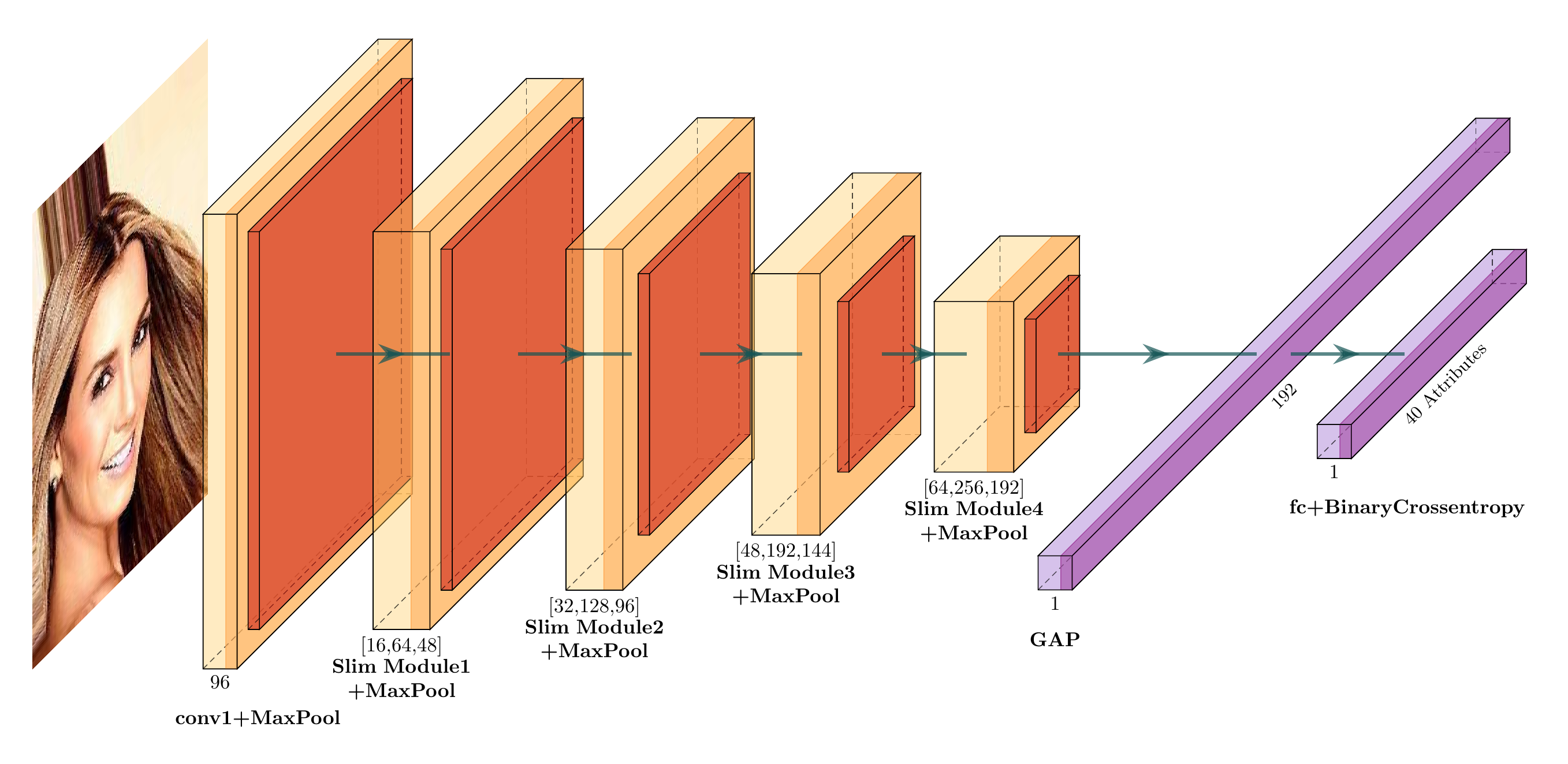}
\caption{\textbf{Full Architecture of our Slim-CNN.} The network consists of 4 stacked \textit{Slim Modules}. Each module is followed by a max-pooling layer. For each \textit{Slim Module}, the value of filter dimensions for different layers are shown in brackets in the following order: Squeeze, Expand and last 3x3 depthwise separable layer.}
\label{fig:FireNet}
\end{figure*}

\subsection{Network}
We use \textit{Slim Modules} as the building blocks of our deep neural network \textit{Slim-CNN}. The proposed micro-architecture yields better discriminative features, while reducing the number of required parameters due to the presence of the SSE blocks within the micro-architecture. Four Slim Modules are stacked together to construct the feature extractor portion of the network. Each Slim Module is followed by max-pooling layer. The feature extractor is followed by global average pooling layer whose output is shared by a fully-connected classification layer of size 40 (.i.e. the total number of attributes). A Global Average Pooling layer was compared to fully-connected layers of size 512 and 1024. The Global Average Layer is preferred as it both reduces the number of parameters and improves the performance of the network. In this work, the \textit{FilterCount} values are set, in order, to [16, 32, 48, 64] for the four Slim Module comprising the Slim-CNN as shown in Figure \ref{fig:FireNet}. In this manner, we can design the complete network without exhaustively trying different configurations for each convolutional layer. The design details of Slim-Net is shown in Table \ref{SlimNetTable}.

Since the task at hand is a binary, multi-label classification problem, we use a sigmoid cross-entropy loss function for each attribute during training. The loss function for each attribute is defined as: 
\begin{equation}
    \mathcal{L}_{attr} = - ~ (y)\log(p) + (1-y)\log(1-p)
\end{equation}
where \textit{y} is the attribute label and \textit{p} is the prediction from corresponding attribute label. 

\section{Experiments and Results}
\subsection{Implementation Details}
All experiments in this work were run using GeForce GTX 1080 Ti on Intel Core i7 with 31 GiB of memory using the Keras Python Deep Learning Library with a TensorFlow backend. For training the network, we used the ADAM optimizer with a learning rate of 0.0001. The network was trained with glorot initialization  \cite{glorot}. The batch-size is set to 64 and the depthwise separable convolution layers in the Slim modules apply L2 regularization(0.0001).  

\begin{table*}[!h]
\footnotesize
\caption{Accuracy, memory and parameter count comparison with state-of-the-art methods. Memory usage reported here is on-disk space used by the model.} \vspace{2pt}
\label{tab:MethodsComp}
\centering
\begin{tabular}{cccccccc}
\hline
\textbf{Attributes}          & \textbf{\begin{tabular}[c]{@{}c@{}}Liu\\  \cite{liu}\end{tabular}} & \textbf{\begin{tabular}[c]{@{}c@{}}MOON\\  \cite{moon}\end{tabular}} & \textbf{\begin{tabular}[c]{@{}c@{}}Hand\\  \cite{hand}\end{tabular}} & \textbf{\begin{tabular}[c]{@{}c@{}}DMTL\\  \cite{han}\end{tabular}}     & \textbf{\begin{tabular}[c]{@{}c@{}}AFFACT\\  \cite{Affact}\end{tabular}}   & \textbf{\begin{tabular}[c]{@{}c@{}}PSE\\  \cite{PSEE}\end{tabular}}     & \textbf{Slim (this work)} \\ \hline 
\textbf{5 O'Clock Shadow}    & 91           & 94.03         & 94.51         & 95                & 94.87             & 94.85            & 94.75             \\ 
\textbf{Arched Eyebrows}     & 79           & 82.26         & 83.42         & 86                & 84.21             & 83.75            & 83.2              \\ 
\textbf{Attractive}          & 81           & 81.67         & 83.06         & 85                & 82.86             & 82.95            & 82.85             \\ 
\textbf{Bags Under Eyes}     & 79           & 84.92         & 84.92         & 99                & 85.31             & 85.35            & 85.33             \\ 
\textbf{Bald}                & 98           & 98.77         & 98.9          & 99                & 99.09             & 98.99            & 98.84             \\ 
\textbf{Bangs}               & 95           & 95.8          & 96.05         & 96                & 96.1              & 95.54            & 96.23             \\ 
\textbf{Big Lips}            & 68           & 71.48         & 71.47         & 88                & 72.7              & 71.9             & 71.44             \\ 
\textbf{Big Nose}            & 78           & 84            & 84.53         & 92                & 84.38             & 84.09            & 84.28             \\ 
\textbf{Black Hair}          & 88           & 89.4          & 89.78         & 85                & 90.36             & 89.76            & 90.13             \\ 
\textbf{Blond Hair}          & 95           & 95.86         & 96.01         & 91                & 96.2              & 95.5             & 95.55             \\ 
\textbf{Blurry}              & 84           & 95.67         & 96.17         & 96                & 96.37             & 96.29            & 95.42             \\ 
\textbf{Brown Hair}          & 80           & 89.38         & 89.15         & 96                & 88.4              & 88.66            & 88.5              \\ 
\textbf{Bushy Eyebrows}      & 90           & 92.62         & 92.84         & 85                & 92.46             & 91.77            & 92.79             \\ 
\textbf{Chubby}              & 91           & 95.44         & 95.67         & 97                & 95.74             & 95.94            & 95.57             \\ 
\textbf{Double Chin}         & 92           & 96.32         & 96.32         & 99                & 96.52             & 96.45            & 96.3              \\ 
\textbf{Eyeglasses}          & 99           & 99.47         & 99.63         & 99                & 99.61             & 99.65            & 99.64             \\ 
\textbf{Goatee}              & 95           & 97.04         & 97.24         & 98                & 97.5              & 97.15            & 97.55             \\ 
\textbf{Gray Hair}           & 97           & 98.1          & 98.2          & 96                & 98.31             & 98.23            & 97.76             \\ 
\textbf{Heavy Makeup}        & 90           & 90.99         & 91.55         & 92                & 92.1              & 90.39            & 91.71             \\ 
\textbf{High Cheekbones}     & 88           & 87.01         & 87.58         & 88                & 87.77             & 87.12            & 87.8              \\ 
\textbf{Male}                & 98           & 98.1          & 98.17         & 98                & 98.5              & 98.21            & 98.12             \\ 
\textbf{Mouth Slightly Open} & 92           & 93.54         & 93.74         & 94                & 94.06             & 93.85            & 93.79             \\ 
\textbf{Mustache}            & 95           & 96.82         & 96.88         & 97                & 97.11             & 96.85            & 97                \\ 
\textbf{Narrow Eyes}         & 81           & 86.52         & 87.23         & 90                & 87.69             & 87.35            & 87.7              \\ 
\textbf{No Beard}            & 95           & 95.58         & 96.05         & 97                & 96.45             & 96.22            & 96.26             \\ 
\textbf{Oval Face}           & 66           & 75.73         & 75.84         & 78                & 77.41             & 73.98            & 75.43             \\ 
\textbf{Pale Skin}           & 91           & 97            & 97.05         & 97                & 97.05             & 97.04            & 97.18             \\ 
\textbf{Pointy Nose}         & 72           & 76.46         & 77.47         & 78                & 76.9              & 77.3             & 77.13             \\ 
\textbf{Receding Hairline}   & 89           & 93.56         & 93.81         & 94                & 93.67             & 93.55            & 93.1              \\ 
\textbf{Rosy Cheeks}         & 90           & 94.82         & 95.16         & 96                & 95.16             & 95.08            & 95.1              \\ 
\textbf{Sideburns}           & 96           & 97.59         & 97.85         & 98                & 97.84             & 97.87            & 97.73             \\ 
\textbf{Smiling}             & 92           & 92.6          & 92.73         & 94                & 92.96             & 91               & 92.93             \\ 
\textbf{Straight Hair}       & 73           & 82.26         & 83.58         & 85                & 85.26             & 84.93            & 83.14             \\ 
\textbf{Wavy Hair}           & 80           & 82.47         & 83.91         & 87                & 86.25             & 85.35            & 83.77             \\ 
\textbf{Wearing Earrings}    & 82           & 89.6          & 90.43         & 90                & 90.98             & 90.26            & 90.38             \\ 
\textbf{Wearing Hat}         & 99           & 98.95         & 99.05         & 99                & 99.1              & 99               & 99.07             \\ 
\textbf{Wearing Lipstick}    & 93           & 93.93         & 94.11         & 93                & 93.96             & 93.41            & 94.18             \\ 
\textbf{Wearing Necklace}    & 71           & 87.04         & 86.63         & 89                & 89.27             & 88.01            & 87.12             \\ 
\textbf{Wearing Necktie}     & 93           & 96.63         & 96.51         & 97                & 97.29             & 97.03            & 96.7              \\ 
\textbf{Young}               & 87           & 88.08         & 88.48         & 90                & 88.98             & 88.67            & 88.14             \\ \hline
\textbf{Average}             & 87           & 90.94         & 91.27         & \textbf{92.1}     & 91.67             & 91.23            & \textbf{91.24}    \\ 
\textbf{Memory}             &  -          & 457MB          & 63.8MB         & 260MB     & 98.2MB             &  -           & \textbf{7.9MB}    \\ 
\textbf{\#Parameters}        &\textgreater{}100M              & 136M          & 15M           & 65M & 26M & 62M & \textbf{0.6M}     \\ \hline
\end{tabular}
\end{table*}

\subsection{Dataset}
In this work, we use the CelebA dataset, which consists of over 200,000 celebrity images where each image has been annotated with 40 different attributes. The dataset covers large variations in pose, background, and illumination which makes predicting attributes extremely challenging. The images in the dataset are split in the following fashion: 160,000 images for training, 20,000 images for validation, and 20,000 images for final testing. Some of the images from the CelebA dataset are shown in Fig \ref{fig:AttributeDefinition}. The CelebA database is an imbalanced set which further illustrates the difficulty of the task. 

\subsection{Attribute Prediction Results}
In this section, we describe the results of many experiments comparing our Slim-CNN against other methods and micro-architectures. 

\subsubsection{Comparison with other methods}
Table \ref{tab:MethodsComp} shows how our Slim-Module inspired deep neural network performs as compared to other methods in terms of both accuracy for each of the 40 attributes and overall model size. As can be seen, our Slim-CNN is very competitive with state-of-the-art methods, while only using a fraction of the parameters of its competitors. Some of these methods use computationally expensive network architecture as the backbone of their proposed solutions. For example, PSE \cite{PSEE}, DMTL\cite{han}, and Liu \cite{liu} are inspired by the parameter intensive Alex-Net (60M parameters) while AFFACT \cite{Affact} is built on top of a ResNet-50 architecture with more than 25M parameters. The reduction in number of parameters of our proposed network is staggering, making it ideal for mobile and embedded applications. We report the on-disk memory needed by the networks for several methods. The proposed Slim-Net is extremely light and require significantly less memory space as compared to other methods, while still achieving high accuracy.

\subsubsection{Comparison of Slim-CNN with other Micro-Architecture-Based DNNs}
Here, we compare our Slim-CNN with other micro-architecture-based DNNs such as Squeeze-Net. To construct the Squeeze-Net DNN used in our experiments, we removed the 3x3 depthwise separable convolution layer after the second SSE Block and the skip connection in each Slim Module. The filter sizes for all the convolutional layers in the resulting Fire modules remain the same as those of the Slim Module. We also construct two different CNNs for comparing the Slim Module with inception modules and residual blocks. These CNNs were constructed by replacing Slim Modules with their corresponding micro-architectures i.e. inception modules and residual blocks. For a fair comparison, we try to keep the configurations of these CNNs consistent with Slim-CNN. For the inception inspired CNN, the number of filters in the 3x3 kernels of the inception module is the same as the number of filters in the 3x3 kernel in the expand layer of SSE blocks of the Slim Module. Similarly for the residual block CNN, the number of filters for the 1x1 and 3x3 kernels in the residual block is twice that of the 1x1 kernel in the squeeze layer and the 3x3 kernel in the expand layer of SSE Block. We tested different configurations for these two CNNs and reported the best results here. We provide more details on the configurations for residual block CNN and inception module CNN in the supplementary material. Table \ref{tab:Comp_MA} shows that Slim-CNN performs better than inception CNN and residual CNN despite having fewer parameters than both. This shows that the Slim Module can efficiently generate good feature representation for faces. Slim-Net improves on the Fire module\cite{SqueezeNet} based DNN results by over 11\% which shows that the addition of a 3x3 kernel after the SSE Blocks and the skip-connection to micro-architecture yields better feature representations.

\subsubsection{Varying the kernel size of the layer after SSE Block}

The Slim Module uses 3x3 depthwise separable convolution layer after the two SSE Blocks. We opted to use a kernel of this size for the final layer of the Slim Module after running several experiments on different kernel sizes. The final layer configurations that were tested are: (i) a single 3x3 depthwise separable convolution layer (ii) a single 5x5 depthwise separable convolution layer (iii) a single 7x7 depthwise separable convolution layer and (iv) two 3x3 depthwise separable convolution layers in series. Table: \ref{tab:Comp_Kernel} shows nearly identical performance from each configuration, suggesting that increasing the size of kernels does not necessarily increase performance. As such, the single 3x3 kernel, having the fewest parameters, was chosen as the configuration for the final layer of the module. This choice was made to be consistent with our design goal of minimizing the parameter space. 

\section{Conclusion}
 We proposed an efficient micro-architecture Slim Module for designing deep neural networks to be used in mobile and embedded applications. Our Slim-CNN has shown comparable performance to the state-of-the-art methods while significantly reducing the number of parameters. Specifically, Slim-CNN achieves an average accuracy of 91.24\%  with only 600K parameters. The next smallest state-of-the-art architecture \cite{hand} has more than 25 times the number of parameters at 16M. We verify the effectiveness of the Slim Module for the face attribute task by comparing it with other well-known micro-architectures in literature. The Slim Module micro-architecture outperforms the Inception module, Residual block, and Fire-Module based SqueezeNet \cite{SqueezeNet}. Moreover, the incredibly low parameter footprint of Slim-CNN makes it very well suited to resource limited environments.

\begin{table}[t]
\footnotesize
\caption{Comparing Performance of Slim-Net with other micro-architectures based DNN} \vspace{2pt}
\label{tab:Comp_MA}
\centering
\begin{tabular}{cccc}
\hline
Models         & Parameters       & Memory         & Accuracy         \\ \hline 
Fire Module\cite{SqueezeNet}    & 764,968                   & 9.8MB                   & 80.10\%                   \\ 
Inception      & 1,224,288                 & 15.5MB                  & 90.62\%                   \\ 
Residual Block & 1,905,096                 & 23.5MB                  & 90.67\%                   \\ 
Slim Module    & \textit{\textbf{570,600}} & \textit{\textbf{7.5MB}} & \textit{\textbf{91.24\%}} \\ \hline
\end{tabular}
\end{table}

\begin{table}[t]
\footnotesize
\caption{Effect of varying the kernel size of the layer after SSE Block.} \vspace{2pt}
\label{tab:Comp_Kernel}
\centering
\begin{tabular}{ccccc}
\hline
Kernel Size & 3x3    &5x5 & 7x7 & 3x3+3x3 \\ \hline
Accuracy    & \textit{\textbf{91.24\%}} & 91.18\%      & 91.16\%      & 91.20\%          \\ 
Parameters  & \textit{\textbf{570,600}} & 591,080      & 615,848      & 680,320          \\ \hline
\end{tabular}
\end{table}


\bibliography{egbib}

\begin{thebibliography}{23}
\providecommand{\natexlab}[1]{#1}
\providecommand{\url}[1]{\texttt{#1}}
\expandafter\ifx\csname urlstyle\endcsname\relax
  \providecommand{\doi}[1]{doi: #1}\else
  \providecommand{\doi}{doi: \begingroup \urlstyle{rm}\Url}\fi

\bibitem[Chen et~al.(2016)Chen, Papandreou, Kokkinos, Murphy, and
  Yuille]{DeepLab}
Liang{-}Chieh Chen, George Papandreou, Iasonas Kokkinos, Kevin Murphy, and
  Alan~L. Yuille.
\newblock Deeplab: Semantic image segmentation with deep convolutional nets,
  atrous convolution, and fully connected crfs.
\newblock \emph{CoRR}, abs/1606.00915, 2016.

\bibitem[Glorot and Bengio(2010)]{glorot}
Xavier Glorot and Yoshua Bengio.
\newblock Understanding the difficulty of training deep feedforward neural
  networks.
\newblock In \emph{Proceedings of the thirteenth international conference on
  artificial intelligence and statistics}, pages 249--256, 2010.

\bibitem[G{\"{u}}nther et~al.(2016)G{\"{u}}nther, Rozsa, and Boult]{Affact}
Manuel G{\"{u}}nther, Andras Rozsa, and Terrance~E. Boult.
\newblock {AFFACT} - alignment free facial attribute classification technique.
\newblock \emph{CoRR}, abs/1611.06158, 2016.
\newblock URL \url{http://arxiv.org/abs/1611.06158}.

\bibitem[Han et~al.(2018)Han, Jain, Wang, Shan, and Chen]{han}
Hu~Han, Anil~K Jain, Fang Wang, Shiguang Shan, and Xilin Chen.
\newblock Heterogeneous face attribute estimation: A deep multi-task learning
  approach.
\newblock \emph{IEEE transactions on pattern analysis and machine
  intelligence}, 40\penalty0 (11):\penalty0 2597--2609, 2018.

\bibitem[Hand and Chellappa(2017)]{hand}
Emily~M Hand and Rama Chellappa.
\newblock Attributes for improved attributes: A multi-task network utilizing
  implicit and explicit relationships for facial attribute classification.
\newblock In \emph{AAAI}, pages 4068--4074, 2017.

\bibitem[He et~al.(2016)He, Zhang, Ren, and Sun]{he}
Kaiming He, Xiangyu Zhang, Shaoqing Ren, and Jian Sun.
\newblock Deep residual learning for image recognition.
\newblock In \emph{Proceedings of the IEEE conference on computer vision and
  pattern recognition}, pages 770--778, 2016.

\bibitem[He et~al.(2017)He, Gkioxari, Doll{\'a}r, and Girshick]{mask}
Kaiming He, Georgia Gkioxari, Piotr Doll{\'a}r, and Ross Girshick.
\newblock Mask r-cnn.
\newblock In \emph{Computer Vision (ICCV), 2017 IEEE International Conference
  on}, pages 2980--2988. IEEE, 2017.

\bibitem[Howard et~al.(2017)Howard, Zhu, Chen, Kalenichenko, Wang, Weyand,
  Andreetto, and Adam]{MobileNet}
Andrew~G. Howard, Menglong Zhu, Bo~Chen, Dmitry Kalenichenko, Weijun Wang,
  Tobias Weyand, Marco Andreetto, and Hartwig Adam.
\newblock Mobilenets: Efficient convolutional neural networks for mobile vision
  applications.
\newblock \emph{CoRR}, abs/1704.04861, 2017.

\bibitem[Iandola et~al.(2016)Iandola, Moskewicz, Ashraf, Han, Dally, and
  Keutzer]{SqueezeNet}
Forrest~N. Iandola, Matthew~W. Moskewicz, Khalid Ashraf, Song Han, William~J.
  Dally, and Kurt Keutzer.
\newblock Squeezenet: Alexnet-level accuracy with 50x fewer parameters and
  {\textless}1mb model size.
\newblock \emph{CoRR}, abs/1602.07360, 2016.

\bibitem[Krizhevsky et~al.(2014)Krizhevsky, Sutskever, and Hinton]{AlexNet}
Alex Krizhevsky, I~Sutskever, and G~Hinton.
\newblock Imagenet classification with deep convolutional neural.
\newblock In \emph{Neural Information Processing Systems}, pages 1--9, 2014.

\bibitem[Kumar et~al.(2009)Kumar, Berg, Belhumeur, and Nayar]{kumar}
Neeraj Kumar, Alexander~C Berg, Peter~N Belhumeur, and Shree~K Nayar.
\newblock Attribute and simile classifiers for face verification.
\newblock In \emph{Computer Vision, 2009 IEEE 12th International Conference
  on}, pages 365--372. IEEE, 2009.

\bibitem[Lin et~al.(2013)Lin, Chen, and Yan]{nin}
Min Lin, Qiang Chen, and Shuicheng Yan.
\newblock Network in network.
\newblock \emph{arXiv preprint arXiv:1312.4400}, 2013.

\bibitem[Liu et~al.(2016)Liu, Anguelov, Erhan, Szegedy, Reed, Fu, and
  Berg]{SSD}
Wei Liu, Dragomir Anguelov, Dumitru Erhan, Christian Szegedy, Scott~E. Reed,
  Cheng{-}Yang Fu, and Alexander~C. Berg.
\newblock {SSD:} single shot multibox detector.
\newblock In \emph{Computer Vision - {ECCV} 2016 - 14th European Conference,
  Amsterdam, The Netherlands, October 11-14, 2016, Proceedings, Part {I}},
  pages 21--37, 2016.

\bibitem[Liu et~al.(2015)Liu, Luo, Wang, and Tang]{liu}
Ziwei Liu, Ping Luo, Xiaogang Wang, and Xiaoou Tang.
\newblock Deep learning face attributes in the wild.
\newblock In \emph{Proceedings of the IEEE International Conference on Computer
  Vision}, pages 3730--3738, 2015.

\bibitem[Rudd et~al.(2016)Rudd, G{\"u}nther, and Boult]{moon}
Ethan~M Rudd, Manuel G{\"u}nther, and Terrance~E Boult.
\newblock Moon: A mixed objective optimization network for the recognition of
  facial attributes.
\newblock In \emph{European Conference on Computer Vision}, pages 19--35.
  Springer, 2016.

\bibitem[Simonyan and Zisserman(2014)]{VGG}
Karen Simonyan and Andrew Zisserman.
\newblock Very deep convolutional networks for large-scale image recognition.
\newblock \emph{CoRR}, abs/1409.1556, 2014.

\bibitem[Srivastava et~al.(2015)Srivastava, Greff, and Schmidhuber]{highway}
Rupesh~Kumar Srivastava, Klaus Greff, and J{\"{u}}rgen Schmidhuber.
\newblock Training very deep networks.
\newblock In \emph{Advances in Neural Information Processing Systems 28: Annual
  Conference on Neural Information Processing Systems 2015, December 7-12,
  2015, Montreal, Quebec, Canada}, pages 2377--2385, 2015.

\bibitem[Szegedy et~al.(2015)Szegedy, Liu, Jia, Sermanet, Reed, Anguelov,
  Erhan, Vanhoucke, and Rabinovich]{szegedy}
Christian Szegedy, Wei Liu, Yangqing Jia, Pierre Sermanet, Scott Reed, Dragomir
  Anguelov, Dumitru Erhan, Vincent Vanhoucke, and Andrew Rabinovich.
\newblock Going deeper with convolutions.
\newblock In \emph{Proceedings of the IEEE conference on computer vision and
  pattern recognition}, pages 1--9, 2015.

\bibitem[Xie et~al.(2017)Xie, Girshick, Doll{\'a}r, Tu, and He]{xie}
Saining Xie, Ross Girshick, Piotr Doll{\'a}r, Zhuowen Tu, and Kaiming He.
\newblock Aggregated residual transformations for deep neural networks.
\newblock In \emph{Computer Vision and Pattern Recognition (CVPR), 2017 IEEE
  Conference on}, pages 5987--5995. IEEE, 2017.

\bibitem[Yi et~al.(2014)Yi, Lei, Liao, and Li]{Casia}
Dong Yi, Zhen Lei, Shengcai Liao, and Stan~Z Li.
\newblock Learning face representation from scratch.
\newblock \emph{arXiv preprint arXiv:1411.7923}, 2014.

\bibitem[Zhang et~al.(2018{\natexlab{a}})Zhang, Zhou, Lin, and Sun]{ShuffleNet}
Xiangyu Zhang, Xinyu Zhou, Mengxiao Lin, and Jian Sun.
\newblock Shufflenet: An extremely efficient convolutional neural network for
  mobile devices.
\newblock In \emph{2018 {IEEE} Conference on Computer Vision and Pattern
  Recognition, {CVPR} 2018, Salt Lake City, UT, USA, June 18-22, 2018}, pages
  6848--6856, 2018{\natexlab{a}}.

\bibitem[Zhang et~al.(2018{\natexlab{b}})Zhang, Shen, Sun, and Li]{PSEE}
Yan Zhang, Wanxia Shen, Li~Sun, and Qingli Li.
\newblock Position-squeeze and excitation block for facial attribute analysis.
\newblock In \emph{British Machine Vision Conference 2018, {BMVC} 2018,
  Northumbria University, Newcastle, UK, September 3-6, 2018}, page 279,
  2018{\natexlab{b}}.

\bibitem[Zhong et~al.(2016)Zhong, Sullivan, and Li]{zhong}
Yang Zhong, Josephine Sullivan, and Haibo Li.
\newblock Leveraging mid-level deep representations for predicting face
  attributes in the wild.
\newblock In \emph{Image Processing (ICIP), 2016 IEEE International Conference
  on}, pages 3239--3243. IEEE, 2016.

\end{thebibliography}
\end{document}